\documentclass[10pt,twocolumn,letterpaper]{article}

\usepackage{cvpr}
\usepackage{times}
\usepackage{epsfig}
\usepackage{graphicx}
\usepackage{amsmath}
\usepackage{amssymb}

\newcommand{\bigfigurewidth}{17.0cm}
\newcommand{\figurewidth}{8.0cm}

\newcommand{\colwidthA}{1.7cm}
\newcommand{\colwidthB}{1.8cm}

\usepackage[breaklinks=true,bookmarks=false]{hyperref}

\cvprfinalcopy 

\def\cvprPaperID{258} 

\begin{document}

\title{InterActive: Inter-Layer Activeness Propagation}

\author{Lingxi Xie\textsuperscript{1$\dagger$}
\thanks{This work was partly done when Lingxi Xie and Liang Zheng were interns at MSR.
They contributed equally.
This work was supported by NIH Grant 5R01EY022247-03, ONR N00014-12-1-0883 and ARO grant W911NF-15-1-0290,
Faculty Research Gift Awards by NEC Labs of America and Blippar, and NSFC 61429201.
We thank John Flynn, Xuan Dong, Jianyu Wang, Junhua Mao and Zhuotun Zhu for discussions.}\quad
\quad Liang Zheng\textsuperscript{2$\dagger$}\quad Jingdong Wang\textsuperscript{3}
\quad Alan Yuille\textsuperscript{4}\quad Qi Tian\textsuperscript{5}\\
\textsuperscript{1,4}Department of Statistics, University of California, Los Angeles, Los Angeles, CA, USA\\
\textsuperscript{3}Microsoft Research, Beijing, China\\
\textsuperscript{4}Departments of Cognitive Science and Computer Science, Johns Hopkins University, Baltimore, MD, USA\\
\textsuperscript{2,5}Department of Computer Science, University of Texas at San Antonio, San Antonio, TX, USA\\
\textsuperscript{1}{\tt\small 198808xc@gmail.com}\quad
\textsuperscript{2}{\tt\small liangzheng06@gmail.com}\\
\textsuperscript{3}{\tt\small jingdw@microsoft.com}\quad
\textsuperscript{4}{\tt\small alan.l.yuille@gmail.com}\quad
\textsuperscript{5}{\tt\small qitian@cs.utsa.edu}
}

\maketitle

\begin{abstract}
An increasing number of computer vision tasks can be tackled with deep features,
which are the intermediate outputs of a pre-trained Convolutional Neural Network.
Despite the astonishing performance, deep features extracted from low-level neurons are still below satisfaction,
arguably because they cannot access the spatial context contained in the higher layers.
In this paper, we present {\bf InterActive}, a novel algorithm which computes the {\bf activeness} of neurons and network connections.
Activeness is propagated through a neural network in a top-down manner,
carrying high-level context and improving the descriptive power of low-level and mid-level neurons.
Visualization indicates that neuron activeness can be interpreted as spatial-weighted neuron responses.
We achieve state-of-the-art classification performance on a wide range of image datasets.
\end{abstract}

\section{Introduction}
\label{Introduction}

We have witnessed a big revolution in computer vision brought by the deep Convolutional Neural Networks (CNN).
With powerful computational resources and a large amount of labeled training data~\cite{Deng_2009_ImageNet},
a differentiable function for classification is trained~\cite{Krizhevsky_2012_ImageNet}
to capture different levels of visual concepts organized by a hierarchical structure.
A pre-trained deep network is also capable of generating deep features for various tasks,
such as image classification~\cite{Jia_2014_CAFFE}\cite{Donahue_2014_DeCAF},
image retrieval~\cite{Razavian_2014_CNN}\cite{Xie_2015_Image} and object detection~\cite{Girshick_2014_Rich}.

Although deep features outperform conventional image representation models such as Bag-of-Visual-Words (BoVW),
we note that the deep feature extraction process only involves forward propagation:
an image is rescaled into a fixed size, input into a pre-trained network,
and the intermediate neuron responses are summarized as visual features.
As we shall see in Section~\ref{Algorithm:Motivation}, such a method ignores important high-level visual context,
causing both a ``big'' problem and a ``small'' problem (see Figure~\ref{Fig:TwoProblems}).
These problems harm the quality of the deep features, and, consequently, visual recognition accuracy.

In this paper, we present {\bf InterActive},
a novel deep feature extraction algorithm which integrates high-level visual context with low-level neuron responses.
For this, we measure the {\em activeness} of neuron connections for each specified image,
based on the idea that a connection is more important if the network output is more sensitive to it.
We define an unsupervised probabilistic distribution function over the high-level neuron responses,
and compute the {\em score function} (a concept in statistics) with respect to each connection.
Each neuron obtains its {\em activeness} by collecting the activeness of the related connections.
InterActive increases the receptive field size of low-level neurons by allowing the supervision of the high-level neurons.
We interpret neuron activeness in terms of spatial-weighted neuron responses,
and the visualization of neuron weights demonstrates that visually salient regions are detected in an unsupervised manner.
More quantitatively, using the improved InterActive features,
we achieve state-of-the-art image classification performance on several popular benchmarks.

The remainder of this paper is organized as follows.
Section~\ref{RelatedWorks} briefly introduces related works.
The InterActive algorithm is presented in Section~\ref{Algorithm}.
Experiments are shown in Section~\ref{Experiments}, and we conclude this work in Section~\ref{Conclusions}.

\section{Related Works}
\label{RelatedWorks}

Image classification is a fundamental problem in computer vision.
In recent years, researchers have extended the conventional tasks~\cite{Lazebnik_2006_Beyond}\cite{Feifei_2007_Learning}
to fine-grained~\cite{Nilsback_2008_Automated}\cite{Wah_2011_Caltech}\cite{Parkhi_2012_Cats},
and large-scale~\cite{Griffin_2007_Caltech}\cite{Xiao_2010_SUN}\cite{Deng_2009_ImageNet} cases.

The Bag-of-Visual-Words (BoVW) model~\cite{Csurka_2004_Visual} represents each images with a high-dimensional vector.
It typically consists of three stages, {\em i.e.}, descriptor extraction, feature encoding and feature summarization.
Due to the limited descriptive power of raw pixels,
local descriptors such as SIFT~\cite{Lowe_2004_Distinctive} and HOG~\cite{Dalal_2005_Histograms} are extracted.
A visual vocabulary is then built to capture the data distribution in feature space.
Descriptors are thereafter quantized onto the vocabulary as
compact feature vectors~\cite{Yang_2009_Linear}\cite{Wang_2010_Locality}\cite{Perronnin_2010_Improving}\cite{Xie_2014_Spatial},
and summarized as an image-level representation~\cite{Lazebnik_2006_Beyond}\cite{Feng_2011_Geometric}\cite{Zhu_2012_Image}.
These feature vectors are post-processed~\cite{Xie_2015_Simple},
and then fed into a machine learning tool~\cite{Fan_2008_LIBLINEAR}\cite{Boiman_2008_Defense}\cite{Xie_2015_Image} for evaluation.

The Convolutional Neural Network (CNN) serves as a hierarchical model for large-scale visual recognition.
It is based on that a network with enough neurons is able to fit any complicated data distribution.
In past years, neural networks were shown to be effective for simple recognition tasks~\cite{LeCun_1990_Handwritten}.
More recently, the availability of large-scale training data ({\em e.g.}, ImageNet~\cite{Deng_2009_ImageNet}) and powerful GPUs
makes it possible to train deep CNNs~\cite{Krizhevsky_2012_ImageNet} which significantly outperform BoVW models.
A CNN is composed of several stacked layers,
in each of which responses from the previous layer are convoluted and activated by a differentiable function.
Hence, a CNN can be considered as a composite function,
and is trained by back-propagating error signals defined by the difference between supervised and predicted labels at the top level.
Recently, efficient methods were proposed to help CNNs converge faster~\cite{Krizhevsky_2012_ImageNet}
and prevent over-fitting~\cite{Hinton_2012_Improving}\cite{Ioffe_2015_Batch}\cite{Xie_2016_DisturbLabel}.
It is believed that deeper networks produce better recognition results~\cite{Simonyan_2015_Very}\cite{Szegedy_2015_Going}.

The intermediate responses of CNN, or the so-called deep features,
serve as efficient image description~\cite{Donahue_2014_DeCAF}, or a set of latent visual attributes.
They can be used for various vision applications, including image classification~\cite{Jia_2014_CAFFE},
image retrieval~\cite{Razavian_2014_CNN}\cite{Xie_2015_Image},
object detection~\cite{Girshick_2014_Rich}\cite{Girshick_2015_Fast} and object parsing~\cite{Xia_2016_Pose}.
A discussion of how different CNN configurations impact deep feature performance is available in~\cite{Chatfield_2014_Return}.

Visualization is an effective method of understanding CNNs.
In~\cite{Zeiler_2014_Visualizing}, a {\em de-convolutional} operation was designed
to capture visual patterns on different layers of a pre-trained network.
\cite{Simonyan_2014_Deep} and \cite{Cao_2015_Look} show that different sets of neurons are activated
when a network is used for detecting different visual concepts.
The above works are based on a {\em supervised} signal on the output layer.
In this paper, we define an {\em unsupervised} probabilistic distribution function on the high-level neuron responses,
and back-propagate it to obtain the activeness of low-level neurons.
Neuron activeness can also be visualized as spatial weighting maps.
Computing neuron activeness involves
finding the relevant contents on each network layer~\cite{Mnih_2014_Recurrent}\cite{Chorowski_2015_Attention},
and is related to recovering low-level details from high-level visual context~\cite{Long_2015_Fully}.

\section{Inter-Layer Activeness Propagation}
\label{Algorithm}

\subsection{Motivation}
\label{Algorithm:Motivation}

\begin{figure}
\begin{center}
    \includegraphics[width=\figurewidth]{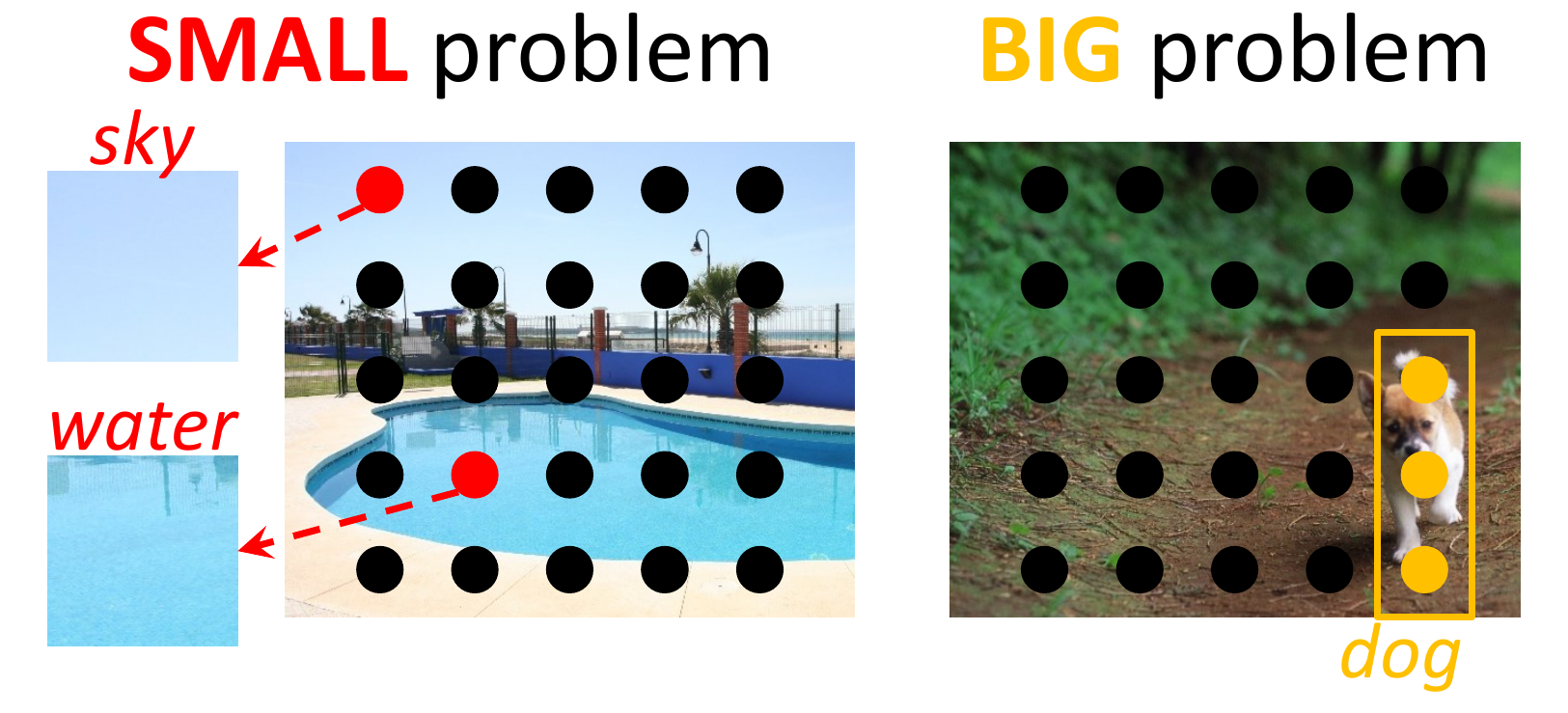}
\end{center}
\caption{
    Examples showing the ``big'' problem and the ``small'' problem of deep feature extraction (best viewed in color PDF).
    Left image (label: {\em swimming pool}):
    the receptive regions of two red neurons are visually similar,
    but correspond to different semantics ({\em sky} vs. {\em water}),
    implying that the receptive field of low-level neurons is often too {\em small} to capture contexts.
    Right image (label: a {\em dog} species):
    only the yellow neurons are located on the object,
    but standard deep features used in classification are pooled over all neurons,
    most of which are irrelevant, suggesting that the pooling region is often too {\em big} compared to the semantic region.
}
\label{Fig:TwoProblems}
\end{figure}

We start with deep features extracted from a pre-trained CNN.
Throughout this paper, we will use the very deep {\bf VGGNet}~\cite{Simonyan_2015_Very} with $19$ convolutional layers.
This produces competitive performance to {\bf GoogLeNet}~\cite{Szegedy_2015_Going},
and outperforms {\bf AlexNet}~\cite{Krizhevsky_2012_ImageNet} significantly.
We also adopt the same notation for layers used in {\bf VGGNet}, {\em e.g.}, {\em conv-3-3}, {\em pool-5} and {\em fc-7}.
All the referred neuron responses are ReLU-processed, {\em i.e.}, negative values are replaced by $0$.

One of the popular deep feature extraction approaches works as follows:
an image is warped (resized) to the same size as the input of a pre-trained network ({\em e.g.} $224\times224$ in {\bf VGGNet}),
then fed into the network, and the responses at an intermediate layer ({\em e.g.}, {\em fc-6}) are used for image representation.
A key observation of~\cite{Simonyan_2015_Very} is that
recognition accuracy is significantly boosted if the input images are not warped.
In what follows, we resize an image, so that the number of pixels is approximately $512^2$,
both width and height are divisible by $32$ (the down-sampling ratio of {\bf VGGNet}), and the aspect ratio is maximally preserved.
Using this setting, we obtain a 3D data cube at each layer (even for {\em fc-6} and {\em fc-7}),
and perform average-pooling or max-pooling to aggregate it as image representation.
We emphasize that such a simple resizing modification gives significant improvement in recognition accuracy.
For example, with features extracted from the {\em fc-6} layer,
the classification accuracy is $83.51\%$, $61.30\%$ and $93.54\%$
on the {\bf Caltech256}, {\bf SUN-397} and {\bf Flower-102} datasets,
whereas features extracted from warped images only report $80.41\%$, $53.06\%$ and $84.89\%$, respectively.
On the {\em pool-5} layer, the numbers are $81.40\%$, $55.22\%$ and $94.70\%$ for un-warped input images,
and $77.46\%$, $48.19\%$ and $86.87\%$ for warped ones, also showing significant improvement.

Compared to the large input image size (approximately $512^2$ pixels),
the receptive field of a neuron on an intermediate layer is much smaller.
For example, a neuron on the {\em pool-4}, {\em pool-5} and {\em fc-6} layers
can {\em see} $124\times124$, $268\times268$ and $460\times460$ pixels on the input image, respectively,
while its effective receptive field is often much smaller~\cite{Zeiler_2014_Visualizing}\cite{Zhou_2015_Object}.
We argue that small receptive fields cause the following problems:
(1) a low-level neuron may not see enough visual context to make prediction,
and (2) there may be many irrelevant neurons which contaminate the image representation.
We name them the ``small'' problem and the ``big'' problem, respectively, as illustrated in Figure~\ref{Fig:TwoProblems}.

Both the above problems can be solved if low-level neurons receive more visual information from higher levels.
In the network training process, this is achieved by error back-propagation,
in which low-level neurons are supervised by high-level neurons to update network weights.
In this section, we present {\bf InterActive},
which is an unsupervised method allowing back-propagating high-level context on the testing stage.
InterActive involves defining a probabilistic distribution function (PDF) on the high-level neuron responses,
and computing the {\em score function} which corresponds to the {\em activeness} of network connections.
As we will see in Section~\ref{Algorithm:Visualization}, this is equivalent to adding spatial weights on low-level neuron responses.

\subsection{The Activeness of Network Connections}
\label{Algorithm:Connections}

Let a deep CNN be a mathematical function ${\mathbf{h}\!\left(\mathbf{X}^{\left(0\right)};\boldsymbol{\Theta}\right)}$,
in which $\mathbf{X}^{\left(0\right)}$ denotes the input image and $\boldsymbol{\Theta}$ the weights over neuron connections.
There are in total $L$ layers,
and the response on the $t$-th layer is $\mathbf{X}^{\left(t\right)}$ (${t}={0}$ indicates the input layer).
In our approach, $\mathbf{X}^{\left(t\right)}$ is a vector of length $W_t\times H_t\times D_t$,
where $W_t$, $H_t$ and $D_t$ denote the width, height and depth (number of channels), respectively.
$x_{w,h,d}^{\left(t\right)}$ is a neuron on the $t$-th layer.
The connections on the $t$-th layer, $\boldsymbol{\theta}^{\left(t\right)}$,
are a matrix of $\left(W_t\times H_t\times D_t\right)\times\left(W_{t+1}\times H_{t+1}\times D_{t+1}\right)$ elements,
where $\theta_{w,h,d,w',h',d'}^{\left(t\right)}$ connects neurons $x_{w,h,d}^{\left(t\right)}$ and $x_{w',h',d'}^{\left(t+1\right)}$.
Let $\mathcal{U}_{w,h,d}^{\left(t\right)}$ be the set of neurons on the $\left(t+1\right)$-st layer
that are connected to $x_{w,h,d}^{\left(t\right)}$,
and $\mathcal{V}_{w',h',d'}^{\left(t+1\right)}$ be the set of neurons on the $t$-th layer
that are connected to $x_{w',h',d'}^{\left(t+1\right)}$.
Hence, the convolution operation can be written as:
\begin{equation}
\label{Eqn:Convolution}
{x_{w',h',d'}^{\left(t+1\right)}}=
    {\sigma\!\left[{\sum_{\left(w,h,d\right)\in\mathcal{V}_{w',h',d'}^{\left(t+1\right)}}}
        x_{w,h,d}^{\left(t\right)}\cdot\theta_{w,h,d,w',h',d'}^{\left(t\right)}+b\right]},
\end{equation}
where ${b}={b_{w',h',d'}^{\left(t+1\right)}}$ is the bias term,
and $\sigma\!\left[\cdot\right]$ is the ReLU activation: ${\sigma\!\left[\cdot\right]}={\max\left(\cdot,0\right)}$.

\begin{figure}
\begin{center}
    \includegraphics[width=\figurewidth]{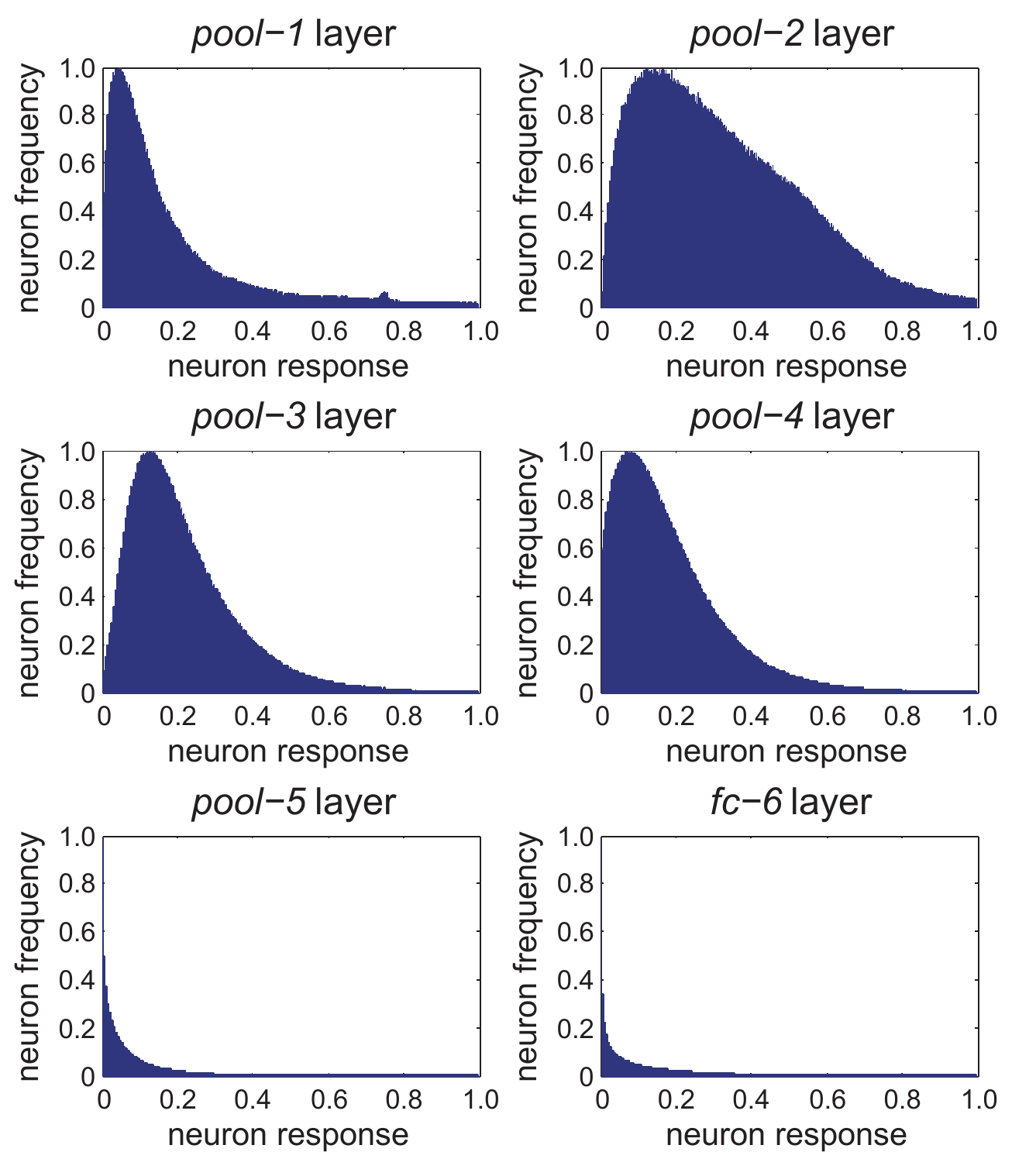}
\end{center}
\caption{
    The statistics of neuron responses on different layers.
    For better visualization, we have filtered all the $0$-responses and normalized the neuron responses and the neuron frequency.
}
\label{Fig:NeuronStatistics}
\end{figure}

We study the PDF on the $T$-th layer $f\!\left(\mathbf{x}^{\left(T\right)}\right)$ by sampling,
where ${\mathbf{x}^{\left(T\right)}}={\left(x_1^{\left(T\right)},\ldots,x_{D_T}^{\left(T\right)}\right)^\top}$
is the averaged neuron response vector over all spatial positions:
\begin{equation}
\label{Eqn:ResponseVector}
{x_d^{\left(T\right)}}={\frac{1}{W_T\times H_T}{\sum_{w=0}^{W_T-1}}{\sum_{w=0}^{H_T-1}}x_{w,h,d}^{\left(T\right)}}.
\end{equation}
We use the {\bf Caltech256} dataset which contains $30607$ natural images to simulate the distribution.
We simply assume that all the $D_T$ elements in $\mathbf{x}^{\left(T\right)}$ are nearly independent,
and summarize all the $30607\times D_T$ elements by 1D histograms shown in Figure~\ref{Fig:NeuronStatistics}.
We can observe that there are typically fewer neurons with large responses.
Therefore, we can assume that the PDF of high-level neurons has the following form:
${f\!\left(\mathbf{x}^{\left(T\right)}\right)}={C_p\cdot\exp\!\left\{-\left\|\mathbf{x}^{\left(T\right)}\right\|_p^p\right\}}$,
where $p$ is the {\em norm} and $C_p$ is the normalization coefficient.

In statistics, the {\em score function} indicates how a likelihood function depends on its parameters.
The score function has been used to produce discriminative features from generative models~\cite{Jaakkola_1999_Exploiting},
{\em e.g.}, as of in Fisher vectors~\cite{Perronnin_2010_Improving}.
It is obtained by computing the gradient of the log-likelihood with respect to the parameters.
Given an image $\mathbf{X}^{\left(0\right)}$,
we compute the intermediate network output $\mathbf{X}^{\left(T\right)}$,
the response vector $\mathbf{x}^{\left(T\right)}$ using~\eqref{Eqn:ResponseVector},
and the likelihood ${f^{\left(T\right)}}\doteq{f\!\left(\mathbf{x}^{\left(T\right)}\right)}$.
Then we compute the score function with respect to $\boldsymbol{\theta}^{\left(t\right)}$
to measure the {\bf activeness} of each network connection in $\boldsymbol{\theta}^{\left(t\right)}$:
\begin{equation}
{\frac{\partial\ln f^{\left(T\right)}}{\partial\boldsymbol{\theta}^{\left(t\right)}}}=
    {\frac{\partial\ln f^{\left(T\right)}}{\partial\mathbf{X}^{\left(t+1\right)}}\cdot
        \frac{\partial\mathbf{X}^{\left(t+1\right)}}{\partial\boldsymbol{\theta}^{\left(t\right)}}},
\end{equation}
where $\mathbf{X}^{\left(t+1\right)}$ is taken as the intermediate term since
it directly depends on $\boldsymbol{\theta}^{\left(t\right)}$.
The two terms on the right-handed side are named the {\em layer-score} and the {\em inter-layer activeness}, respectively.

\subsubsection{The Layer Score}
\label{Algorithm:Connections:LayerScore}

We first compute the layer score $\frac{\partial\ln f^{\left(T\right)}}{\partial\mathbf{X}^{\left(t+1\right)}}$.
From the chain rule of differentiation we have:
\begin{equation}
{\frac{\partial\ln f^{\left(T\right)}}{\partial\mathbf{X}^{\left(t+1\right)}}}=
    {\frac{\partial\ln f^{\left(T\right)}}{\partial\mathbf{X}^{\left(T\right)}}\cdot
        \frac{\partial\mathbf{X}^{\left(T\right)}}{\partial\mathbf{X}^{\left(t+1\right)}}}
\end{equation}
The second term on the right-handed side,
{\em i.e.}, $\frac{\partial\mathbf{X}^{\left(T\right)}}{\partial\mathbf{X}^{\left(t+1\right)}}$,
can be easily derived by network back-propagation as in the training process.
The only difference is that the gradient on the top ($T$-th) layer is defined by
$\frac{\partial\ln f^{\left(T\right)}}{\partial\mathbf{X}^{\left(T\right)}}$.
From $\mathbf{x}^{\left(T\right)}$ defined in~\eqref{Eqn:ResponseVector}
and ${f^{\left(T\right)}}={C_p\cdot\exp\!\left\{-\left\|\mathbf{x}^{\left(T\right)}\right\|_p^p\right\}}$,
we have:
\begin{equation}
\label{Eqn:LayerScoreLp}
{\frac{\partial\ln f^{\left(T\right)}}{\partial\mathbf{X}^{\left(T\right)}}}=
    {-\frac{p}{W_T\times H_T}\cdot\left(\mathbf{x}^{\left(T\right)}\right)^{p-1}\cdot
        \frac{\partial\mathbf{x}^{\left(T\right)}}{\partial\mathbf{X}^{\left(T\right)}}}
\end{equation}
where $\left(\mathbf{X}^{\left(T\right)}\right)^{p-1}$ is the element-wise $\left(p-1\right)$-st power of the vector.
In particular, when ${p}={1}$, the layer score is proportional to an all-one vector $\mathbf{1}^{W_T\times H_T\times D_T}$;
when ${p}={2}$, each of the $W_T\times H_T$ sections is proportional $\mathbf{x}^{\left(T\right)}$.

\subsubsection{The Inter-Layer Activeness}
\label{Algorithm:Connections:InterActive}

Next we compute the inter-layer activeness
$\frac{\partial\mathbf{X}^{\left(t+1\right)}}{\partial\boldsymbol{\theta}^{\left(t\right)}}$.
Consider a single term $\frac{\partial x_{w',h',d'}^{\left(t+1\right)}}{\partial\theta_{w,h,d,w',h',d'}^{\left(t\right)}}$,
direct differentiation of~\eqref{Eqn:Convolution} gives:
\begin{equation}
\label{Eqn:InterLayerGradientReLU}
{\frac{\partial x_{w',h',d'}^{\left(t+1\right)}}{\partial\theta_{w,h,d,w',h',d'}^{\left(t\right)}}}=
    {x_{w,h,d}^{\left(t\right)}\cdot\mathbb{I}_{x_{w',h',d'}^{\left(t+1\right)}>0}\cdot
        \mathbb{I}_{\left(w',h',d'\right)\in\mathcal{U}_{w,h,d}^{\left(t\right)}}},
\end{equation}
where $\mathbb{I}_\cdot$ is the indicator whose value is $1$ when the conditional term is true and $0$ otherwise.

\subsection{The Activeness of Neurons}
\label{Algorithm:Neurons}

With the layer score~\eqref{Eqn:LayerScoreLp} and the inter-layer gradient~\eqref{Eqn:InterLayerGradientReLU},
the score function with respect to $\boldsymbol{\theta}^{\left(t\right)}$ is derived to be:
\begin{equation}
\label{Eqn:ScoreFunction}
{\frac{\partial\ln f^{\left(T\right)}}{\partial\theta_{w,h,d,w',h',d'}^{\left(t\right)}}}=
    {x_{w,h,d}^{\left(t\right)}\cdot\alpha_{w,h,d,w',h',d'}^{\left(t\right)}},
\end{equation}
where ${\alpha_{w,h,d,w',h',d'}^{\left(t\right)}}$
is the {\em importance} of the neuron $x_{w,h,d}^{\left(t\right)}$
to the connection between $x_{w,h,d}^{\left(t\right)}$ and $x_{w',h',d'}^{\left(t+1\right)}$:
${\alpha_{w,h,d,w',h',d'}^{\left(t\right)}}\doteq
    {\mathbb{I}_{x_{w',h',d'}^{\left(t+1\right)}>0}\cdot\mathbb{I}_{\left(w',h',d'\right)\in\mathcal{U}_{w,h,d}^{\left(t\right)}}\cdot
        \frac{\partial\ln f^{\left(T\right)}}{\partial x_{w',h',d'}^{\left(t+1\right)}}}$.
Recall that the score function can be used as visual features.
Therefore, we define the {\bf activeness} of each neuron by accumulating the activeness of all the related connections:
\begin{equation}
\label{Eqn:NeuronActiveness}
{\widetilde{x}_{w,h,d}^{\left(t\right)}}=
    {{\sum_{\left(w',h',d'\right)\in\mathcal{U}_{w,h,d}^{\left(t\right)}}}
        \frac{\partial\ln f^{\left(T\right)}}{\partial\theta_{w,h,d,w',h',d'}^{\left(t\right)}}}.
\end{equation}

We summarize ${\widetilde{\mathbf{X}}^{\left(t\right)}}=
    {\left\{\widetilde{x}_{w,h,d}^{\left(t\right)}\right\}^{W_t\times H_t\times D_t}}$ with max-pooling~\eqref{Eqn:ResponseVector},
resulting in a $D_t$-dimensional {\bf InterActive} feature vector $\widetilde{\mathbf{x}}^{\left(t\right)}$.
As we will see in Section~\ref{Experiments:Configurations},
$\widetilde{\mathbf{x}}^{\left(t\right)}$ is a discriminative representation of the input image $\mathbf{X}^{\left(0\right)}$.

The relationship between $T$ and $t$ can be arbitrary, provided it satisfies ${T}\geqslant{t+1}$.
In this paper, we consider two typical settings, {\em i.e.}, ${T}={L}$ ($L$ is the number of layers) and ${T}={t+1}$,
which means that the supervision comes from the final layer ({\em i.e.}, {\em fc-7}) or its direct successor.
We name them the {\em last} and the {\em next} configurations, respectively.

\subsection{Visualization}
\label{Algorithm:Visualization}

\begin{figure*}
\begin{center}
    \includegraphics[width=\bigfigurewidth]{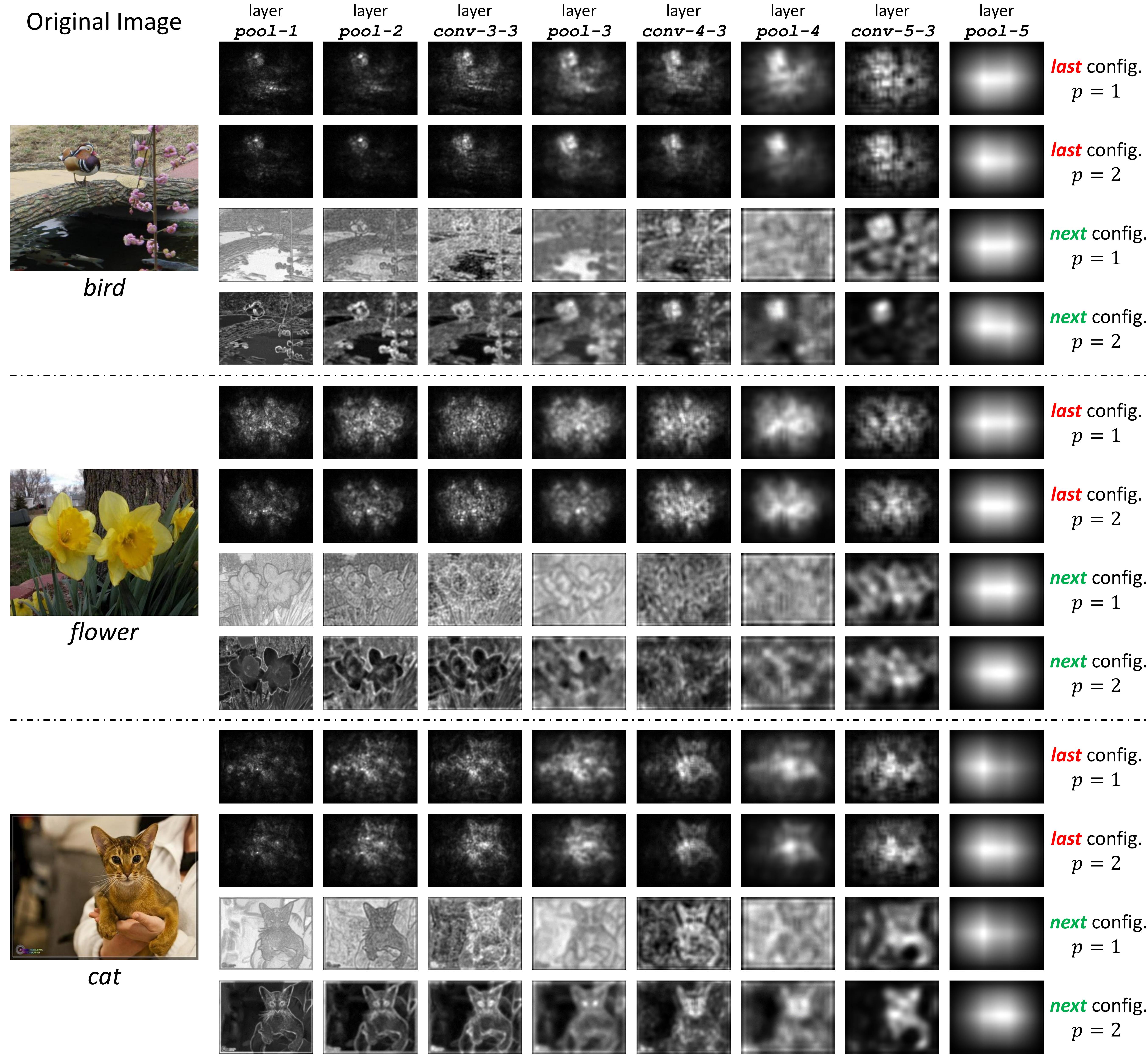}
\end{center}
\caption{
    Typical visualization results of activeness $\widehat{\gamma}_{w,h}^{\left(t\right)}$ with different configurations.
    Neuron weighting maps are resized to the original image size for better visualization.
    Neurons with larger activeness are plotted with higher intensity values (closer to {\em white}).
    Regarding the {\em last} and {\em next} configurations,
    please refer to the texts in Section~\ref{Algorithm:Visualization} for details.
}
\label{Fig:Visualization}
\end{figure*}

Before using the InterActive features for experiments (Section~\ref{Experiments}),
we note that $\widetilde{x}_{w,h,d}^{\left(t\right)}$
is a weighted version of the original neuron response $x_{w,h,d}^{\left(t\right)}$.
The weighting term is:
\begin{equation}
\label{Eqn:SpatialWeighting}
{\gamma_{w,h,d}^{\left(t\right)}}=
    {{\sum_{\left(w',h',d'\right)\in\mathcal{U}_{w,h,d}^{\left(t\right)}}}\alpha_{w,h,d,w',h',d'}^{\left(t\right)}}.
\end{equation}
It counts the activated ({\em i.e.}, $x_{w',h',d'}^{\left(t+1\right)}>0$) neurons on the $\left(t+1\right)$-st layer,
with the importance $\frac{\partial\ln f^{\left(T\right)}}{\partial x_{w',h',d'}^{\left(t+1\right)}}$,
which is supervised by a higher level (the $T$-th layer).

We visualize the weighting term $\gamma_{w,h,d}^{\left(t\right)}$
on the 2D image plane by defining ${\widehat{\gamma}_{w,h}^{\left(t\right)}}={{\sum_d}\gamma_{w,h,d}^{\left(t\right)}}$.
The weighting map is then resized to the original image size.
Representative results are shown in Figure~\ref{Fig:Visualization}.
We observe that spatial weighting weakly captures the interest regions,
although the network is pre-trained using an independent set ({\em i.e.}, {\bf ImageNet}~\cite{Deng_2009_ImageNet}).
Here, we discuss how different parameters affect the weighting terms.

First, activeness measures the contribution of each neuron to higher-level visual outputs.
For a low-level neuron, if the supervision comes from the {\em next} layer, its receptive field is not significantly enlarged
({\em e.g.}, a neuron on the {\em pool-1} receives information from the {\em next} layer
to increase the receptive field from $6\times6$ to $18\times18$).
Therefore, it is more likely that local high-contrast regions becomes more activated,
and the weighting map looks like boundary detection results.
As $t$ increases, neurons have larger receptive fields and capture less local details,
thus the weighting map is more similar to saliency detection results.

Second, the {\em last} and {\em next} configurations make a big difference in activeness,
especially for the low-level and mid-level neurons.
Supervised by the top layer, the {\em last} configuration generates stable weighting maps,
with the high-weight regions corresponding to the salient objects on the image.
However, the output of the {\em next} configuration is quite sensitive to small noises,
and sometimes the background regions even receive more attention than the semantic objects.
As we will see in experiments (Section~\ref{Experiments:Configurations}),
the {\em last} configuration consistently produces higher recognition accuracy on the low-level and mid-level features.

We also compare different {\em norms}, {\em i.e.}, ${p}={1}$ vs. ${p}={2}$.
When ${p}={2}$, spatial weighting rewards neurons with high responses more heavily,
and the high-activeness regions become more concentrated.
In general, $p$ reflects the extent that we assume high-response neurons are more important.
Although other $p$ values can be used,
we believe that ${p}={1}$ and ${p}={2}$ are sufficient to illustrate the difference and produce good performance.
We also test ${p}\rightarrow{+\infty}$, which only considers the neuron with the maximal response,
but the performance is inferior to that using ${p}={1}$ and ${p}={2}$.

\subsection{Comparison to Related Works}
\label{Algorithm:Comparison}

Although both InterActive and network training involve gradient back-propagation, they are propagating different information.
In the training process, a {\em supervised} loss function is defined by the difference between ground-truth and predicted outputs.
In deep feature extraction, however, there is no ground-truth, so we define an {\em unsupervised} loss using the score function.
Both methods lead to propagating high-level visual context through the network to enhance the descriptive power of low-level neurons.

Although our method and~\cite{Zeiler_2014_Visualizing} share similar ideas, they are quite different.
We focus on generating better image description, while~\cite{Zeiler_2014_Visualizing} focuses on visualizing the network;
we can visualize back-propagated neuron activeness, while~\cite{Zeiler_2014_Visualizing} visualizes neuron responses;
we back-propagate the activeness of all neurons, while~\cite{Zeiler_2014_Visualizing} only chooses the neuron with maximal response;
our method is unsupervised, while~\cite{Zeiler_2014_Visualizing} is supervised (by ``guessing'' the label).
Being unsupervised, InterActive can be generalized to many more classification problems with a different set of image classes.

In another work on object detection~\cite{Cao_2015_Look}, the neural network is {\em told} a visual concept,
and the supervised signal is back-propagated to find the most relevant neurons.
InterActive performs detection in an implicit, unsupervised manner, making it feasible to be applied to image classification.

\section{Experiments}
\label{Experiments}

\begin{table*}
\begin{center}
\begin{tabular}{|l|l|r|p{\colwidthA}|p{\colwidthA}|p{\colwidthA}|p{\colwidthA}|p{\colwidthA}|p{\colwidthA}|}
\hline
Layer          & Model           & Dims   &
    {\bf Caltech256} & {\bf Indoor-67}  & {\bf SUN-397}    & {\bf Pet-37}     & {\bf Flower-102} & {\bf Bird-200}   \\
\hline\hline
{\em pool-1}   & Orig., AVG      & $  64$ &
    $11.12$          & $19.96$          & $ 8.52$          & $12.09$          & $29.36$          & $ 5.10$          \\
{\em pool-1}   & Orig., MAX      & $  64$ &
    $ 8.77$          & $16.82$          & $ 7.27$          & $14.83$          & $27.95$          & $ 7.81$          \\
{\em pool-1}   & Next, ${p}={1}$ & $  64$ &
    $11.01$          & $19.97$          & $ 8.62$          & $11.60$          & $29.11$          & $ 4.95$          \\
{\em pool-1}   & Next, ${p}={2}$ & $  64$ &
    $11.26$          & $19.71$          & $ 8.92$          & $12.38$          & $31.07$          & $ 5.30$          \\
{\em pool-1}   & Last, ${p}={1}$ & $  64$ &
    $12.93$          & $20.83$          & $ 9.83$          & $20.64$          & $32.93$          & $ 8.55$          \\
{\em pool-1}   & Last, ${p}={2}$ & $  64$ &
    $\mathbf{13.14}$ & $\mathbf{21.10}$ & $\mathbf{10.02}$ & $\mathbf{21.19}$ & $\mathbf{33.58}$ & $\mathbf{ 9.01}$ \\
\hline
{\em pool-2}   & Orig., AVG      & $ 128$ &
    $21.03$          & $31.12$          & $18.63$          & $20.49$          & $45.77$          & $ 8.30$          \\
{\em pool-2}   & Orig., MAX      & $ 128$ &
    $19.47$          & $28.29$          & $16.05$          & $24.60$          & $43.39$          & $11.28$          \\
{\em pool-2}   & Next, ${p}={1}$ & $ 128$ &
    $20.98$          & $30.93$          & $18.59$          & $19.89$          & $45.62$          & $ 8.01$          \\
{\em pool-2}   & Next, ${p}={2}$ & $ 128$ &
    $20.65$          & $30.95$          & $19.01$          & $21.18$          & $48.27$          & $ 9.60$          \\
{\em pool-2}   & Last, ${p}={1}$ & $ 128$ &
    $25.84$          & $33.24$          & $20.25$          & $37.29$          & $53.72$          & $18.52$          \\
{\em pool-2}   & Last, ${p}={2}$ & $ 128$ &
    $\mathbf{26.20}$ & $\mathbf{33.47}$ & $\mathbf{20.50}$ & $\mathbf{38.42}$ & $\mathbf{54.22}$ & $\mathbf{19.43}$ \\
\hline
{\em conv-3-3} & Orig., AVG      & $ 256$ &
    $26.44$          & $36.42$          & $22.73$          & $27.78$          & $49.70$          & $10.47$          \\
{\em conv-3-3} & Orig., MAX      & $ 256$ &
    $24.18$          & $33.27$          & $19.71$          & $31.43$          & $48.02$          & $13.85$          \\
{\em conv-3-3} & Next, ${p}={1}$ & $ 256$ &
    $27.29$          & $36.97$          & $22.84$          & $28.89$          & $50.62$          & $10.93$          \\
{\em conv-3-3} & Next, ${p}={2}$ & $ 256$ &
    $27.62$          & $37.36$          & $23.41$          & $30.38$          & $54.06$          & $12.73$          \\
{\em conv-3-3} & Last, ${p}={1}$ & $ 256$ &
    $34.50$          & $39.40$          & $25.84$          & $49.41$          & $60.53$          & $24.21$          \\
{\em conv-3-3} & Last, ${p}={2}$ & $ 256$ &
    $\mathbf{35.29}$ & $\mathbf{39.68}$ & $\mathbf{26.02}$ & $\mathbf{50.57}$ & $\mathbf{61.06}$ & $\mathbf{25.27}$ \\
\hline
{\em pool-3}   & Orig., AVG      & $ 256$ &
    $29.17$          & $37.98$          & $23.59$          & $29.88$          & $52.44$          & $11.00$          \\
{\em pool-3}   & Orig., MAX      & $ 256$ &
    $26.53$          & $34.65$          & $20.83$          & $33.68$          & $50.93$          & $13.66$          \\
{\em pool-3}   & Next, ${p}={1}$ & $ 256$ &
    $29.09$          & $38.12$          & $24.05$          & $30.08$          & $52.26$          & $10.89$          \\
{\em pool-3}   & Next, ${p}={2}$ & $ 256$ &
    $29.55$          & $38.61$          & $24.31$          & $31.98$          & $55.06$          & $12.65$          \\
{\em pool-3}   & Last, ${p}={1}$ & $ 256$ &
    $36.96$          & $41.02$          & $26.73$          & $50.91$          & $62.41$          & $24.58$          \\
{\em pool-3}   & Last, ${p}={2}$ & $ 256$ &
    $\mathbf{37.40}$ & $\mathbf{41.45}$ & $\mathbf{27.22}$ & $\mathbf{51.96}$ & $\mathbf{63.06}$ & $\mathbf{25.47}$ \\
\hline
{\em conv-4-3} & Orig., AVG      & $ 512$ &
    $49.62$          & $59.66$          & $42.03$          & $55.57$          & $76.98$          & $21.45$          \\
{\em conv-4-3} & Orig., MAX      & $ 512$ &
    $47.73$          & $55.83$          & $40.10$          & $59.40$          & $75.72$          & $23.39$          \\
{\em conv-4-3} & Next, ${p}={1}$ & $ 512$ &
    $51.83$          & $60.37$          & $43.59$          & $59.29$          & $78.54$          & $25.01$          \\
{\em conv-4-3} & Next, ${p}={2}$ & $ 512$ &
    $53.52$          & $60.65$          & $44.17$          & $63.40$          & $80.48$          & $31.07$          \\
{\em conv-4-3} & Last, ${p}={1}$ & $ 512$ &
    $61.62$          & $62.45$          & $45.43$          & $75.29$          & $85.91$          & $52.26$          \\
{\em conv-4-3} & Last, ${p}={2}$ & $ 512$ &
    $\mathbf{61.98}$ & $\mathbf{62.74}$ & $\mathbf{45.87}$ & $\mathbf{77.61}$ & $\mathbf{86.08}$ & $\mathbf{54.12}$ \\
\hline
{\em pool-4}   & Orig., AVG      & $ 512$ &
    $60.39$          & $66.49$          & $49.73$          & $66.76$          & $85.56$          & $28.56$          \\
{\em pool-4}   & Orig., MAX      & $ 512$ &
    $57.92$          & $62.96$          & $47.29$          & $69.23$          & $84.39$          & $30.01$          \\
{\em pool-4}   & Next, ${p}={1}$ & $ 512$ &
    $60.59$          & $66.48$          & $49.55$          & $66.28$          & $85.68$          & $28.40$          \\
{\em pool-4}   & Next, ${p}={2}$ & $ 512$ &
    $62.06$          & $66.94$          & $50.01$          & $72.40$          & $87.36$          & $37.49$          \\
{\em pool-4}   & Last, ${p}={1}$ & $ 512$ &
    $68.20$          & $67.20$          & $51.04$          & $81.04$          & $91.22$          & $57.41$          \\
{\em pool-4}   & Last, ${p}={2}$ & $ 512$ &
    $\mathbf{68.60}$ & $\mathbf{67.40}$ & $\mathbf{51.30}$ & $\mathbf{82.56}$ & $\mathbf{92.00}$ & $\mathbf{59.25}$ \\
\hline
{\em conv-5-3} & Orig., AVG      & $ 512$ &
    $77.40$          & $74.66$          & $59.47$          & $88.36$          & $94.03$          & $55.44$          \\
{\em conv-5-3} & Orig., MAX      & $ 512$ &
    $75.93$          & $71.38$          & $57.03$          & $87.10$          & $91.30$          & $55.19$          \\
{\em conv-5-3} & Next, ${p}={1}$ & $ 512$ &
    $80.31$          & $\mathbf{74.80}$ & $59.63$          & $90.29$          & $94.84$          & $67.64$          \\
{\em conv-5-3} & Next, ${p}={2}$ & $ 512$ &
    $80.73$          & $74.52$          & $\mathbf{59.74}$ & $\mathbf{91.56}$ & $95.16$          & $\mathbf{73.14}$ \\
{\em conv-5-3} & Last, ${p}={1}$ & $ 512$ &
    $80.77$          & $73.68$          & $59.10$          & $90.73$          & $95.40$          & $69.32$          \\
{\em conv-5-3} & Last, ${p}={2}$ & $ 512$ &
    $\mathbf{80.84}$ & $73.58$          & $58.96$          & $91.19$          & $\mathbf{95.70}$ & $69.75$          \\
\hline
{\em pool-5}   & Orig., AVG      & $ 512$ &
    $81.40$          & $\mathbf{74.93}$ & $\mathbf{55.22}$ & $91.78$          & $94.70$          & $69.72$          \\
{\em pool-5}   & Orig., MAX      & $ 512$ &
    $79.61$          & $71.88$          & $54.04$          & $89.43$          & $90.01$          & $68.52$          \\
{\em pool-5}   & Next, ${p}={1}$ & $ 512$ &
    $81.50$          & $72.70$          & $53.83$          & $92.01$          & $95.41$          & $71.96$          \\
{\em pool-5}   & Next, ${p}={2}$ & $ 512$ &
    $81.58$          & $72.63$          & $53.57$          & $\mathbf{92.30}$ & $95.40$          & $\mathbf{73.21}$ \\
{\em pool-5}   & Last, ${p}={1}$ & $ 512$ &
    $81.60$          & $72.58$          & $53.93$          & $92.20$          & $\mathbf{95.43}$ & $72.47$          \\
{\em pool-5}   & Last, ${p}={2}$ & $ 512$ &
    $\mathbf{81.68}$ & $72.68$          & $53.79$          & $92.18$          & $95.41$          & $72.51$          \\
\hline
{\em fc-6}     & Orig., AVG      & $4096$ &
    $83.51$          & $\mathbf{75.52}$ & $\mathbf{61.30}$ & $93.08$          & $\mathbf{93.54}$ & $\mathbf{71.69}$ \\
{\em fc-6}     & Orig., MAX      & $4096$ &
    $83.59$          & $74.47$          & $59.39$          & $93.07$          & $93.20$          & $71.03$          \\
{\em fc-6}     & Last, ${p}={1}$ & $4096$ &
    $83.44$          & $75.48$          & $61.28$          & $92.84$          & $93.40$          & $70.26$          \\
{\em fc-6}     & Last, ${p}={2}$ & $4096$ &
    $\mathbf{83.61}$ & $75.50$          & $61.19$          & $\mathbf{93.10}$ & $93.45$          & $71.60$          \\
\hline
\end{tabular}
\caption{
    Classification accuracy ($\%$) comparison among different configurations.
    {\bf Bold} numbers indicate the best performance in each group ({\em i.e.}, same dataset, same layer).
    For {\em fc-6}, the {\em next} and {\em last} layers are the same
    (see the texts in Section~\ref{Algorithm:Visualization} for details).
}
\label{Tab:Models}
\end{center}
\end{table*}

\begin{table*}
\begin{center}
\begin{tabular}{|l|p{\colwidthB}|p{\colwidthB}|p{\colwidthB}|p{\colwidthB}|p{\colwidthB}|p{\colwidthB}|}
\hline
Model                                              &
    {\bf Caltech256} & {\bf Indoor-67}  & {\bf SUN-397}    & {\bf Pet-37}     & {\bf Flower-102} & {\bf Bird-200}   \\
\hline\hline
Murray {\em et.al.}~\cite{Murray_2014_Generalized} &
    $-$              & $-$              & $-$              & $56.8 $          & $84.6 $          & $33.3 $          \\
Kobayashi {\em et.al.}~\cite{Kobayashi_2015_Three} &
    $58.3 $          & $64.8 $          & $-$              & $-$              & $-$              & $30.0$           \\
Liu {\em et.al.}~\cite{Liu_2015_Novel}             &
    $75.47$          & $59.12$          & $-$              & $-$              & $-$              & $-$              \\
Xie {\em et.al.}~\cite{Xie_2015_RIDE}              &
    $60.25$          & $64.93$          & $50.12$          & $63.49$          & $86.45$          & $50.81$          \\
\hline
Chatfield {\em et.al}~\cite{Chatfield_2014_Return} &
    $77.61$          & $-$              & $-$              & $-$              & $-$              & $-$              \\
Donahue {\em et.al}~\cite{Donahue_2014_DeCAF}      &
    $-$              & $-$              & $40.94$          & $-$              & $-$              & $64.96$          \\
Razavian {\em et.al.}~\cite{Razavian_2014_CNN}     &
    $-$              & $69.0 $          & $-$              & $-$              & $86.8 $          & $61.8 $          \\
Zeiler {\em et.al}~\cite{Zeiler_2014_Visualizing}  &
    $74.2 $          & $-$              & $-$              & $-$              & $-$              & $-$              \\
Zhou {\em et.al.}~\cite{Zhou_2014_Learning}        &
    $-$              & $69.0 $          & $54.3 $          & $-$              & $-$              & $-$              \\
Qian {\em et.al.}~\cite{Qian_2015_Fine}            &
    $-$              & $-$              & $-$              & $81.18$          & $89.45$          & $67.86$          \\
Xie {\em et.al.}~\cite{Xie_2015_Image}             &
    $-$              & $70.13$          & $54.87$          & $90.03$          & $86.82$          & $62.02$          \\
\hline
Ours (Orig., AVG)      &
    $84.02$          & $78.02$          & $62.30$          & $93.02$          & $95.70$          & $73.35$          \\
Ours (Orig., MAX)      &
    $84.38$          & $77.32$          & $61.87$          & $93.20$          & $95.98$          & $74.76$          \\
Ours (Next, ${p}={1}$) &
    $84.43$          & $78.01$          & $62.26$          & $92.91$          & $96.02$          & $74.37$          \\
Ours (Next, ${p}={2}$) &
    $84.64$          & $78.23$          & $62.50$          & $93.22$          & $96.26$          & $74.61$          \\
Ours (Last, ${p}={1}$) &
    $84.94$          & $78.40$          & $62.69$          & $93.40$          & $96.35$          & $75.47$          \\
Ours (Last, ${p}={2}$) &
    $\mathbf{85.06}$ & $\mathbf{78.65}$ & $\mathbf{62.97}$ & $\mathbf{93.45}$ & $\mathbf{96.40}$ & $\mathbf{75.62}$ \\
\hline
\end{tabular}
\caption{
    Accuracy ($\%$) comparison with recent works (published after 2014) without (above) and with (middle) using deep features.
    We use the concatenated feature vectors from all the $9$ layers used in Table~\ref{Tab:Models}.
    For the {\bf Bird-200} dataset, most competitors use extra information (bounding boxes and/or detected parts) but we do not.
    With bounding boxes, we achieve higher accuracy: $\mathbf{77.53\%}$.
    See texts for details.
}
\label{Tab:Comparison}
\end{center}
\end{table*}

\subsection{Datasets and Settings}
\label{Experiments:DatasetsSettings}

We evaluate InterActive on six popular image classification datasets.
For generic object recognition,
we use the {\bf Caltech256}~\cite{Griffin_2007_Caltech} ($30607$ images, $257$ classes, $60$ training samples for each class) dataset.
For scene recognition,
we use the MIT {\bf Indoor-67}~\cite{Quattoni_2009_Recognizing} ($15620$ images, $67$ classes, $80$ training samples per class)
and the {\bf SUN-397}~\cite{Xiao_2010_SUN} ($108754$ images, $397$ classes, $50$ training samples per class) datasets.
For fine-grained object recognition,
we use the Oxford {\bf Pet-37}~\cite{Parkhi_2012_Cats} ($7390$ images, $37$ classes, $100$ training samples per class),
the Oxford {\bf Flower-102}~\cite{Nilsback_2008_Automated} ($8189$ images, $102$ classes, $20$ training samples per class)
and the Caltech-UCSD {\bf Bird-200}~\cite{Wah_2011_Caltech} ($11788$ images, $200$ classes, $30$ training samples per class) datasets.

We use the $19$-layer {\bf VGGNet}~\cite{Simonyan_2015_Very} (pre-trained on {\bf ImageNet}) for deep features extraction.
We use the model provided by the MatConvNet library~\cite{Vedaldi_2015_MatConvNet} without fine-tuning.
Its down-sampling rate is $32$, caused by the five max-pooling layers.
As described in Section~\ref{Algorithm:Motivation}, we maximally preserve the aspect ratio of the input image,
constrain the width and height divisible by $32$, and the number of pixels is approximately $512^2$.
The InterActive feature vectors are $\ell_2$-normalized and sent to LIBLINEAR~\cite{Fan_2008_LIBLINEAR},
a scalable SVM implementation, with the slacking parameter $C$ fixed as $10$.

\subsection{InterActive Configurations}
\label{Experiments:Configurations}

We evaluate the InterActive features extracted from different layers,
using different {\em norms} $p$, and either the {\em last} or {\em next} configuration
(please refer to Section~\ref{Algorithm:Visualization} and Figure~\ref{Fig:Visualization}).
We also compare InterActive with the original deep features with average-pooling or max-pooling.
Classification results are summarized in Table~\ref{Tab:Models}.

We first observe the low-level and mid-level layers (from {\em pool-1} to {\em pool-4}).
InterActive with the {\em last} configuration consistently outperforms the original deep features.
Sometimes, the accuracy gain is very significant
({\em e.g.}, more than $30\%$ on {\em conv-4-3} and {\em pool4} for {\em bird} recognition),
showing that InterActive improves image representation by letting the low-level and mid-level neurons receive high-level context.
Although these layers often produce low accuracy,
the improvement contributes when multi-level features are combined (see Table~\ref{Tab:Comparison}).
Regarding the {\em norm}, ${p}={2}$ always works better than ${p}={1}$.
Recalling from~\eqref{Eqn:LayerScoreLp} that ${p}={2}$ better rewards high-response neurons,
we conclude that high-response neurons are indeed more important.

On the high-level neurons ({\em i.e.}, {\em pool-5} and {\em fc-6}),
the advantage of InterActive vanishes in scene classification, and the original average-pooled features produce the best accuracy.
Therefore, it is more likely that all the high-level neurons are equally important for scene understanding.
On object recognition tasks, the advantage also becomes much smaller,
since InterActive only provides limited increase on high-level neurons' receptive field.

The intermediate output of the $t$-th layer can be considered as a bunch of $D_t$-dimensional visual descriptors.
Possible choices of feature aggregation include average-pooling and max-pooling.
If each image region approximately contributes equally (such as in scene recognition), average-pooling produces higher accuracy,
however in the case that semantic objects are quite small (such as on the {\bf Bird-200} dataset), max-pooling works better.
InterActive computes neuron activeness in an unsupervised manner,
which provides a soft weighting scheme, or a tradeoff between max-pooling and average-pooling.
By detecting interesting regions automatically, it often produces higher accuracy than both max-pooling and average-pooling.

\subsection{Comparison to the State-of-the-Arts}
\label{Experiments:Comparison}

We compare InterActive with several recent works in Table~\ref{Tab:Comparison}.
These algorithms also extract features from statistics-based methods, and use machine learning tools for classification.
We concatenate the feature vectors of all $9$ layers in Table~\ref{Tab:Models} as a $6848$-dimensional vector.
Apart from the {\bf Bird-200} dataset, the reported accuracy is the highest, to the best of our knowledge.
Although the accuracy gain over baseline is relatively small ({\em e.g.}, $0.43\%$ in {\bf Pet-37}),
we emphasize that the baseline accuracy is already very high, thanks to the improved deep feature extraction strategy.
Therefore, the improvement of InterActive is not so small as it seems.
On the other hand, recognition rates are consistently boosted with InterActive, without requiring extra information,
which demonstrates that deep features can be intrinsically improved when neuron activeness is considered.

On the {\bf Bird-200} dataset, it is very important to detect the position and/or compositional parts
of the objects~\cite{Chai_2013_Symbiotic}\cite{Gavves_2013_Fine}\cite{Zhang_2014_Part}\cite{Xiao_2015_Application},
otherwise heavy computation is required to achieve good performance~\cite{Xie_2015_Image}.
InterActive implicitly finds the semantic object regions, leading to competitive $75.62\%$ accuracy.
If the bounding box of each object is provided (as in~\cite{Zhang_2014_Part} and~\cite{Lin_2015_Deep}),
the original and InterActive features produce $76.95\%$ and $77.53\%$ accuracy, respectively.
Using bounding boxes provides $3.60\%$ and $1.91\%$ accuracy gain on original and InterActive features, respectively.
InterActive significantly reduces the gap with implicit object detection.
$77.53\%$ is lower than $80.26\%$ in~\cite{Lin_2015_Deep} and $82.8\%$ in~\cite{Krause_2015_Fine},
both of which require fine-tuning the network and R-CNN part detection~\cite{Girshick_2014_Rich} while InterActive does not.
We believe that InterActive can cooperate with these strategies.

\subsection{ImageNet Experiments}
\label{Experiments:ImageNet}

We report results on {\bf ILSVRC2012}, a subset of {\bf ImageNet} which contains $1000$ categories.
We use the pre-trained {\bf VGGNet} models and the same image cropping techniques as in~\cite{Simonyan_2015_Very}.
The baseline validation error rates on the $16$-layer model,
the $19$-layer model and the combined model are $7.1\%$, $7.0\%$ and $6.7\%$,
respectively (slightly better than~\cite{Simonyan_2015_Very}).
We apply InterActive to update the neuron responses on the second-to-last layer ({\em fc-7}) and
forward-propagate them to re-compute the classification scores ({\em fc-8}).
The error rates are decreased to $6.8\%$, $6.7\%$ and $6.5\%$, respectively.
The improvement is significant given that the baseline is already high and our method is very simple.

In the future, we will explore the use of InterActive on some challenging datasets,
such as the {\bf PASCAL-VOC} dataset and the {\bf Microsoft COCO} dataset~\cite{Lin_2014_Microsoft}.
We thank the anonymous reviewers for this valuable suggestion.

\section{Conclusions}
\label{Conclusions}

In this paper, we present InterActive, a novel algorithm for deep feature extraction.
We define a probabilistic distribution function on the high-level neuron responses,
and back-propagate the score function through the network to compute the {\em activeness} of each network connection and each neuron.
We reveal that high-level visual context carries rich information to enhance low-level and mid-level feature representation.
The output of our algorithm is the activeness of each neuron, or a weighted version of the original neuron response.
InterActive improves visual feature representation,
and achieves the state-of-the-art performance on several popular image classification benchmarks.

InterActive can be applied to many more vision tasks.
On the one hand, with the {\em last} configuration, neuron activeness provides strong clues for saliency detection.
On the other hand, with the {\em next} configuration on a low-level layer,
neuron activeness can be used to detect local high-contrast regions, which may correspond to edges or boundaries.
All these possibilities are left for future research.

{\small
\bibliographystyle{ieee}
\bibliography{egbib}

\begin{thebibliography}{10}\itemsep=-1pt

\bibitem{Boiman_2008_Defense}
O.~Boiman, E.~Shechtman, and M.~Irani.
\newblock {In Defense of Nearest-Neighbor Based Image Classification}.
\newblock {\em Computer Vision and Pattern Recognition}, 2008.

\bibitem{Cao_2015_Look}
C.~Cao, X.~Liu, Y.~Yang, Y.~Yu, J.~Wang, Z.~Wang, Y.~Huang, L.~Wang, C.~Huang,
  and W.~Xu.
\newblock {Look and Think Twice: Capturing Top-Down Visual Attention with
  Feedback Convolutional Neural Networks}.
\newblock {\em International Conference on Computer Vision}, 2015.

\bibitem{Chai_2013_Symbiotic}
Y.~Chai, V.~Lempitsky, and A.~Zisserman.
\newblock {Symbiotic Segmentation and Part Localization for Fine-Grained
  Categorization}.
\newblock {\em International Conference on Computer Vision}, 2013.

\bibitem{Chatfield_2014_Return}
K.~Chatfield, K.~Simonyan, A.~Vedaldi, and A.~Zisserman.
\newblock {Return of the Devil in the Details: Delving Deep into Convolutional
  Nets}.
\newblock {\em British Machine Vision Conference}, 2014.

\bibitem{Chorowski_2015_Attention}
J.~Chorowski, D.~Bahdanau, D.~Serdyuk, K.~Cho, and Y.~Bengio.
\newblock {Attention-Based Models for Speech Recognition}.
\newblock {\em Advances in Neural Information Processing Systems}, 2015.

\bibitem{Csurka_2004_Visual}
G.~Csurka, C.~Dance, L.~Fan, J.~Willamowski, and C.~Bray.
\newblock {Visual Categorization with Bags of Keypoints}.
\newblock {\em Workshop on Statistical Learning in Computer Vision, European
  Conference on Computer Vision}, 1(22):1--2, 2004.

\bibitem{Dalal_2005_Histograms}
N.~Dalal and B.~Triggs.
\newblock {Histograms of Oriented Gradients for Human Detection}.
\newblock {\em Computer Vision and Pattern Recognition}, pages 886--893, 2005.

\bibitem{Deng_2009_ImageNet}
J.~Deng, W.~Dong, R.~Socher, L.~Li, K.~Li, and L.~Fei-Fei.
\newblock {ImageNet: A Large-Scale Hierarchical Image Database}.
\newblock {\em Computer Vision and Pattern Recognition}, 2009.

\bibitem{Donahue_2014_DeCAF}
J.~Donahue, Y.~Jia, O.~Vinyals, J.~Hoffman, N.~Zhang, E.~Tzeng, and T.~Darrell.
\newblock {DeCAF: A Deep Convolutional Activation Feature for Generic Visual
  Recognition}.
\newblock {\em International Conference on Machine Learning}, 2014.

\bibitem{Fan_2008_LIBLINEAR}
R.~Fan, K.~Chang, C.~Hsieh, X.~Wang, and C.~Lin.
\newblock {LIBLINEAR: A Library for Large Linear Classification}.
\newblock {\em Journal of Machine Learning Research}, 9:1871--1874, 2008.

\bibitem{Feifei_2007_Learning}
L.~Fei-Fei, R.~Fergus, and P.~Perona.
\newblock {Learning Generative Visual Models from Few Training Examples: An
  Incremental Bayesian Approach Tested on 101 Object Categories}.
\newblock {\em Computer Vision and Image Understanding}, 106(1):59--70, 2007.

\bibitem{Feng_2011_Geometric}
J.~Feng, B.~Ni, Q.~Tian, and S.~Yan.
\newblock {Geometric Lp-norm Feature Pooling for Image Classification}.
\newblock {\em Computer Vision and Pattern Recognition}, 2011.

\bibitem{Gavves_2013_Fine}
E.~Gavves, B.~Fernando, C.~Snoek, A.~Smeulders, and T.~Tuytelaars.
\newblock {Fine-Grained Categorization by Alignments}.
\newblock {\em International Conference on Computer Vision}, 2013.

\bibitem{Girshick_2015_Fast}
R.~Girshick.
\newblock {Fast R-CNN}.
\newblock {\em International Conference on Computer Vision}, 2015.

\bibitem{Girshick_2014_Rich}
R.~Girshick, J.~Donahue, T.~Darrell, and J.~Malik.
\newblock {Rich Feature Hierarchies for Accurate Object Detection and Semantic
  Segmentation}.
\newblock {\em Computer Vision and Pattern Recognition}, 2014.

\bibitem{Griffin_2007_Caltech}
G.~Griffin, A.~Holub, and P.~Perona.
\newblock {Caltech-256 Object Category Dataset}.
\newblock {\em Technical Report: CNS-TR-2007-001, Caltech}, 2007.

\bibitem{Hinton_2012_Improving}
G.~Hinton, N.~Srivastava, A.~Krizhevsky, I.~Sutskever, and R.~Salakhutdinov.
\newblock {Improving Neural Networks by Preventing Co-adaptation of Feature
  Detectors}.
\newblock {\em arXiv preprint, arXiv: 1207.0580}, 2012.

\bibitem{Ioffe_2015_Batch}
S.~Ioffe and C.~Szegedy.
\newblock {Batch Normalization: Accelerating Deep Network Training by Reducing
  Internal Covariate Shift}.
\newblock {\em International Conference on Machine Learning}, 2015.

\bibitem{Jaakkola_1999_Exploiting}
T.~Jaakkola, D.~Haussler, et~al.
\newblock {Exploiting Generative Models in Discriminative Classifiers}.
\newblock {\em Advances in Neural Information Processing Systems}, pages
  487--493, 1999.

\bibitem{Jia_2014_CAFFE}
Y.~Jia, E.~Shelhamer, J.~Donahue, S.~Karayev, J.~Long, R.~Girshick,
  S.~Guadarrama, and T.~Darrell.
\newblock {CAFFE: Convolutional Architecture for Fast Feature Embedding}.
\newblock {\em ACM International Conference on Multimedia}, 2014.

\bibitem{Kobayashi_2015_Three}
T.~Kobayashi.
\newblock {Three Viewpoints Toward Exemplar SVM}.
\newblock {\em Computer Vision and Pattern Recognition}, 2015.

\bibitem{Krause_2015_Fine}
J.~Krause, H.~Jin, J.~Yang, and L.~Fei-Fei.
\newblock {Fine-Grained Recognition without Part Annotations}.
\newblock {\em Computer Vision and Pattern Recognition}, 2015.

\bibitem{Krizhevsky_2012_ImageNet}
A.~Krizhevsky, I.~Sutskever, and G.~Hinton.
\newblock {ImageNet Classification with Deep Convolutional Neural Networks}.
\newblock {\em Advances in Neural Information Processing Systems}, 2012.

\bibitem{Lazebnik_2006_Beyond}
S.~Lazebnik, C.~Schmid, and J.~Ponce.
\newblock {Beyond Bags of Features: Spatial Pyramid Matching for Recognizing
  Natural Scene Categories}.
\newblock {\em Computer Vision and Pattern Recognition}, 2006.

\bibitem{LeCun_1990_Handwritten}
Y.~LeCun, J.~Denker, D.~Henderson, R.~Howard, W.~Hubbard, and L.~Jackel.
\newblock {Handwritten Digit Recognition with a Back-Propagation Network}.
\newblock {\em Advances in Neural Information Processing Systems}, 1990.

\bibitem{Lin_2015_Deep}
D.~Lin, X.~Shen, C.~Lu, and J.~Jia.
\newblock {Deep LAC: Deep Localization, Alignment and Classification for
  Fine-grained Recognition}.
\newblock {\em Computer Vision and Pattern Recognition}, 2015.

\bibitem{Lin_2014_Microsoft}
T.-Y. Lin, M.~Maire, S.~Belongie, J.~Hays, P.~Perona, D.~Ramanan,
  P.~Doll{\'a}r, and C.~L. Zitnick.
\newblock {Microsoft COCO: Common Objects in Context}.
\newblock {\em European Conference on Computer Vision}, 2014.

\bibitem{Liu_2015_Novel}
Q.~Liu and C.~Liu.
\newblock {A Novel Locally Linear KNN Model for Visual Recognition}.
\newblock {\em Computer Vision and Pattern Recognition}, 2015.

\bibitem{Long_2015_Fully}
J.~Long, E.~Shelhamer, and T.~Darrell.
\newblock {Fully Convolutional Networks for Semantic Segmentation}.
\newblock {\em Computer Vision and Pattern Recognition}, 2015.

\bibitem{Lowe_2004_Distinctive}
D.~Lowe.
\newblock {Distinctive Image Features from Scale-Invariant Keypoints}.
\newblock {\em International Journal on Computer Vision}, 60(2):91--110, 2004.

\bibitem{Mnih_2014_Recurrent}
V.~Mnih, N.~Heess, and A.~Graves.
\newblock {Recurrent Models of Visual Attention}.
\newblock {\em Advances in Neural Information Processing Systems}, 2014.

\bibitem{Murray_2014_Generalized}
N.~Murray and F.~Perronnin.
\newblock {Generalized Max Pooling}.
\newblock {\em Computer Vision and Pattern Recognition}, 2014.

\bibitem{Nilsback_2008_Automated}
M.~Nilsback and A.~Zisserman.
\newblock {Automated Flower Classification over a Large Number of Classes}.
\newblock {\em Indian Conference on Computer Vision, Graphics \& Image
  Processing}, 2008.

\bibitem{Parkhi_2012_Cats}
O.~Parkhi, A.~Vedaldi, A.~Zisserman, and C.~Jawahar.
\newblock {Cats and Dogs}.
\newblock {\em Computer Vision and Pattern Recognition}, 2012.

\bibitem{Perronnin_2010_Improving}
F.~Perronnin, J.~Sanchez, and T.~Mensink.
\newblock {Improving the Fisher Kernel for Large-scale Image Classification}.
\newblock {\em European Conference on Computer Vision}, 2010.

\bibitem{Qian_2015_Fine}
Q.~Qian, R.~Jin, S.~Zhu, and Y.~Lin.
\newblock {Fine-Grained Visual Categorization via Multi-stage Metric Learning}.
\newblock {\em Computer Vision and Pattern Recognition}, 2015.

\bibitem{Quattoni_2009_Recognizing}
A.~Quattoni and A.~Torralba.
\newblock {Recognizing Indoor Scenes}.
\newblock {\em Computer Vision and Pattern Recognition}, 2009.

\bibitem{Razavian_2014_CNN}
A.~Razavian, H.~Azizpour, J.~Sullivan, and S.~Carlsson.
\newblock {CNN Features off-the-shelf: an Astounding Baseline for Recognition}.
\newblock {\em Computer Vision and Pattern Recognition}, 2014.

\bibitem{Simonyan_2014_Deep}
K.~Simonyan, A.~Vedaldi, and A.~Zisserman.
\newblock {Deep Inside Convolutional Networks: Visualising Image Classification
  Models and Saliency Maps}.
\newblock {\em Workshop of International Conference on Learning
  Representations}, 2014.

\bibitem{Simonyan_2015_Very}
K.~Simonyan and A.~Zisserman.
\newblock {Very Deep Convolutional Networks for Large-Scale Image Recognition}.
\newblock {\em International Conference on Learning Representations}, 2015.

\bibitem{Szegedy_2015_Going}
C.~Szegedy, W.~Liu, Y.~Jia, P.~Sermanet, S.~Reed, D.~Anguelov, D.~Erhan,
  V.~Vanhoucke, and A.~Rabinovich.
\newblock {Going Deeper with Convolutions}.
\newblock {\em Computer Vision and Pattern Recognition}, 2015.

\bibitem{Vedaldi_2015_MatConvNet}
A.~Vedaldi and K.~Lenc.
\newblock {MatConvNet-Convolutional Neural Networks for MATLAB}.
\newblock {\em ACM International Conference on Multimedia}, 2015.

\bibitem{Wah_2011_Caltech}
C.~Wah, S.~Branson, P.~Welinder, P.~Perona, and S.~Belongie.
\newblock {The Caltech-UCSD Birds-200-2011 Dataset}.
\newblock {\em Technical Report: CNS-TR-2011-001, Caltech}, 2011.

\bibitem{Wang_2010_Locality}
J.~Wang, J.~Yang, K.~Yu, F.~Lv, T.~Huang, and Y.~Gong.
\newblock {Locality-Constrained Linear Coding for Image Classification}.
\newblock {\em Computer Vision and Pattern Recognition}, pages 3360--3367,
  2010.

\bibitem{Xia_2016_Pose}
F.~Xia, J.~Zhu, P.~Wang, and A.~Yuille.
\newblock {Pose-Guided Human Parsing by an AND/OR Graph Using Pose-Context
  Features}.
\newblock {\em AAAI Conference on Artificial Intelligence}, 2016.

\bibitem{Xiao_2010_SUN}
J.~Xiao, J.~Hays, K.~Ehinger, A.~Oliva, and A.~Torralba.
\newblock {SUN Database: Large-Scale Scene Recognition from Abbey to Zoo}.
\newblock {\em Computer Vision and Pattern Recognition}, 2010.

\bibitem{Xiao_2015_Application}
T.~Xiao, Y.~Xu, K.~Yang, J.~Zhang, Y.~Peng, and Z.~Zhang.
\newblock {The Application of Two-Level Attention Models in Deep Convolutional
  Neural Network for Fine-Grained Image Classification}.
\newblock {\em Computer Vision and Pattern Recognition}, 2015.

\bibitem{Xie_2015_Image}
L.~Xie, R.~Hong, B.~Zhang, and Q.~Tian.
\newblock {Image Classification and Retrieval are ONE}.
\newblock {\em International Conference on Multimedia Retrieval}, 2015.

\bibitem{Xie_2014_Spatial}
L.~Xie, Q.~Tian, M.~Wang, and B.~Zhang.
\newblock {Spatial Pooling of Heterogeneous Features for Image Classification}.
\newblock {\em IEEE Transactions on Image Processing}, 23(5):1994--2008, 2014.

\bibitem{Xie_2015_Simple}
L.~Xie, Q.~Tian, and B.~Zhang.
\newblock {Simple Techniques Make Sense: Feature Pooling and Normalization for
  Image Classification}.
\newblock {\em IEEE Transactions on Circuits and Systems for Video Technology},
  2015.

\bibitem{Xie_2015_RIDE}
L.~Xie, J.~Wang, W.~Lin, B.~Zhang, and Q.~Tian.
\newblock {RIDE: Reversal Invariant Descriptor Enhancement}.
\newblock {\em International Conference on Computer Vision}, 2015.

\bibitem{Xie_2016_DisturbLabel}
L.~Xie, J.~Wang, Z.~Wei, M.~Wang, and Q.~Tian.
\newblock {DisturbLabel: Regularizing CNN on the Loss Layer}.
\newblock {\em Computer Vision and Pattern Recognition}, 2016.

\bibitem{Yang_2009_Linear}
J.~Yang, K.~Yu, Y.~Gong, and T.~Huang.
\newblock {Linear Spatial Pyramid Matching Using Sparse Coding for Image
  Classification}.
\newblock {\em Computer Vision and Pattern Recognition}, pages 1794--1801,
  2009.

\bibitem{Zeiler_2014_Visualizing}
M.~Zeiler and R.~Fergus.
\newblock {Visualizing and Understanding Convolutional Networks}.
\newblock {\em European Conference on Computer Vision}, 2014.

\bibitem{Zhang_2014_Part}
N.~Zhang, J.~Donahue, R.~Girshick, and T.~Darrell.
\newblock {Part-based R-CNNs for Fine-Grained Category Detection}.
\newblock {\em European Conference on Computer Vision}, 2014.

\bibitem{Zhou_2015_Object}
B.~Zhou, A.~Khosla, A.~Lapedriza, A.~Oliva, and A.~Torralba.
\newblock {Object Detectors Emerge in Deep Scene CNNs}.
\newblock {\em International Conference on Learning Representations}, 2015.

\bibitem{Zhou_2014_Learning}
B.~Zhou, A.~Lapedriza, J.~Xiao, A.~Torralba, and A.~Oliva.
\newblock {Learning Deep Features for Scene Recognition Using Places Database}.
\newblock {\em Advances in Neural Information Processing Systems}, 2014.

\bibitem{Zhu_2012_Image}
J.~Zhu, W.~Zou, X.~Yang, R.~Zhang, Q.~Zhou, and W.~Zhang.
\newblock {Image Classification by Hierarchical Spatial Pooling with Partial
  Least Squares Analysis}.
\newblock {\em British Machine Vision Conference}, 2012.

\end{thebibliography}
}

\end{document}